\documentclass[runningheads]{llncs}
\usepackage{graphicx}

\usepackage{tikz}
\usepackage{comment}
\usepackage{amsmath,amssymb} 
\usepackage{color}

\usepackage[accsupp]{axessibility}  

\usepackage{epsfig}
\usepackage{graphicx}
\usepackage{amsmath}
\usepackage{amssymb}
\usepackage{subfiles}
\usepackage{multirow} 

\usepackage{booktabs}
\usepackage[ruled]{algorithm2e}
\usepackage{xcolor,colortbl}
\usepackage{mmstyle}

\usepackage{cite}
\usepackage{float}
\usepackage{booktabs}
\usepackage{rotating}

\usepackage{enumitem}
\usepackage{transparent}

\usepackage{makecell}
\definecolor{citecolor}{HTML}{0071bc}
\usepackage[pagebackref,breaklinks,colorlinks,citecolor=citecolor]{hyperref}

\definecolor{myred}{RGB}{225,97,78}
\definecolor{SSL}{RGB}{188, 149, 117}
\definecolor{architecture}{RGB}{117, 188, 149}
\definecolor{regularization}{RGB}{117, 156, 188}
\definecolor{data}{RGB}{188, 117, 156}

\def\Bname{OmniBenchmark}

\usepackage{marvosym}
\makeatletter
\def\@fnsymbol#1{\ensuremath{\ifcase#1\or \textsuperscript{~\Letter}\or \ddagger\or
   \mathsection\or \mathparagraph\or \|\or **\or \dagger\dagger
   \or \ddagger\ddagger \else\@ctrerr\fi}}
\makeatother

\begin{document}
\pagestyle{headings}
\mainmatter
\def\ECCVSubNumber{287}  

\title{Benchmarking Omni-Vision Representation \\ through the Lens of Visual Realms} 

\titlerunning{ECCV-22 submission ID \ECCVSubNumber} 
\authorrunning{ECCV-22 submission ID \ECCVSubNumber} 
\author{Anonymous ECCV submission}
\institute{Paper ID \ECCVSubNumber}

\titlerunning{OmniBenchmark}
%
\author{Yuanhan Zhang\inst{1} \and
Zhenfei Yin\inst{2} \and
Jing Shao\inst{2} \and
Ziwei Liu\inst{1}}
\authorrunning{Zhang et al.}
%
\institute{S-Lab, Nanyang Technological University, Singapore \\
\email{\{yuanhan002,ziwei.liu\}@ntu.edu.sg}\\
\and
SenseTime Research \\
\email{\{yinzhenfei,shaojing\}@sensetime.com}}
\maketitle
\setcounter{footnote}{0}

\begin{abstract}

Though impressive performance has been achieved in specific visual realms (\eg faces, dogs, and places), an omni-vision representation generalizing to many natural visual domains is highly desirable.
But, existing benchmarks are biased and inefficient to evaluate the omni-vision representation---these benchmarks either only include several specific realms, or cover most realms at the expense of subsuming numerous datasets that have extensive realm overlapping.
%
In this paper, we propose Omni-Realm Benchmark (\textbf{OmniBenchmark}). It includes 21 realm-wise datasets with 7,372 concepts and 1,074,346 images. Without semantic overlapping, these datasets cover most visual realms comprehensively and meanwhile efficiently.
In addition, we propose a new supervised contrastive learning framework, namely \textbf{Re}lational \textbf{Co}ntrastive learning (\textbf{ReCo}), for a better omni-vision representation.
Beyond pulling two instances from the same concept closer---the typical supervised contrastive learning framework---ReCo also pulls two instances from the same semantic realm closer, encoding the semantic relation between concepts, facilitating omni-vision representation learning.
We benchmark ReCo and other advances in omni-vision representation studies that are different in architectures (from CNNs to transformers) and in learning paradigms (from supervised learning to self-supervised learning) on OmniBenchmark.
We illustrate the superior of ReCo to other supervised contrastive learning methods, and reveal multiple practical observations to facilitate future research. The code and models are available at \url{https://zhangyuanhan-ai.github.io/OmniBenchmark}.

\keywords{Representation Learning, Visual Realm}

\end{abstract}

\section{Introduction}
\label{sec:introduction}

\begin{figure}[t]
\centering
\includegraphics[width=\textwidth]{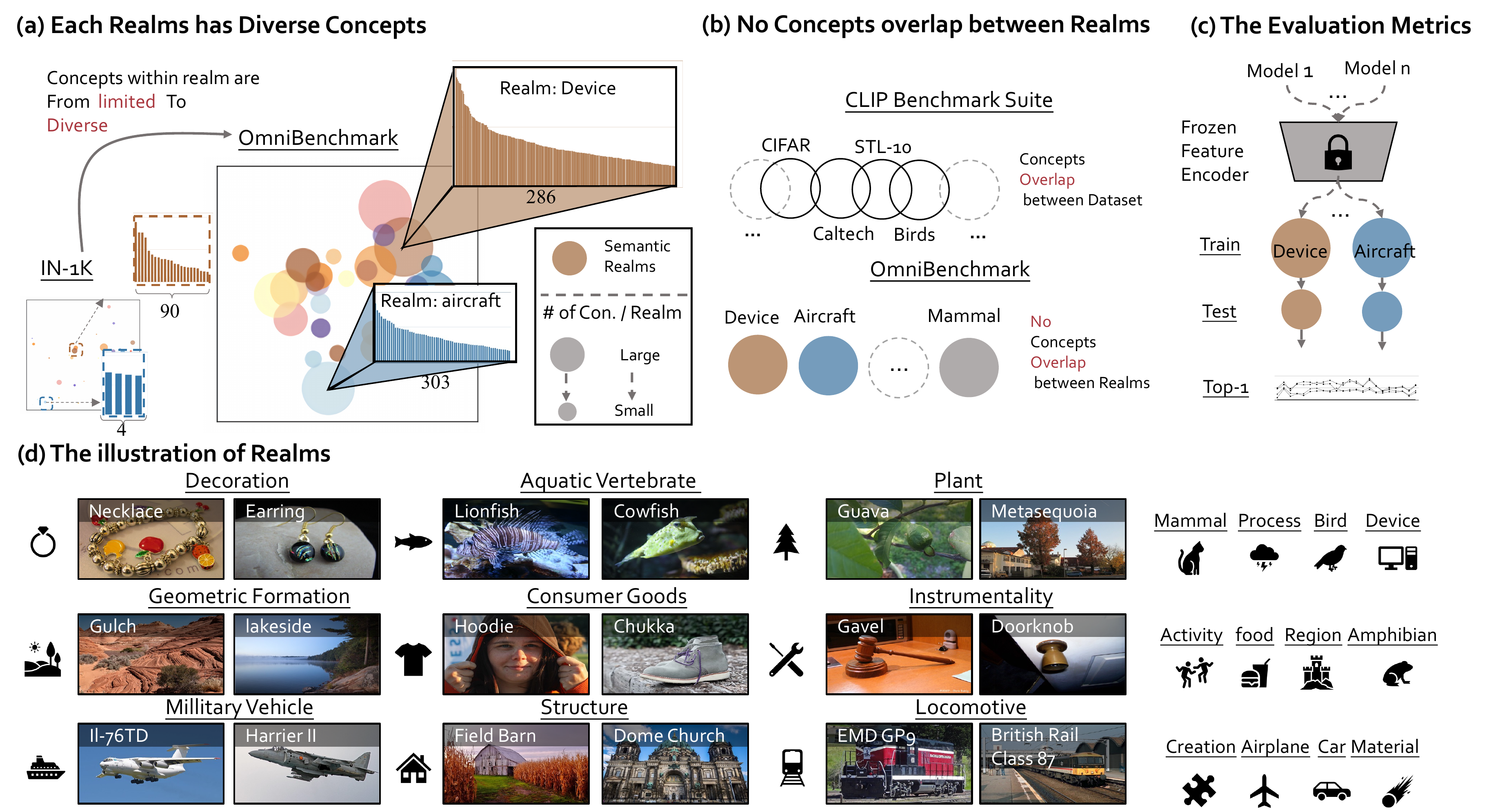}
\caption{\textbf{A overview of \Bname}. (a) Compared with ImageNet-1k, \Bname~covers more realms and annotate more images of each realm. (b) Included datasets in OmniBenchmark have no semantic concepts overlap. (c) OmniBenchmark diagnoses pre-trained models through linear probing. (d) We present 21 semantic realm-wise datasets and the sampled concepts of 9 datasets.
}
\label{fig:fig1}
\end{figure}

%
Large-scale pre-trained models, either trained in a supervised~\cite{SWSL,Re-labeling,resnet} or unsupervised manner~\cite{CLIP,swav,MOCOv1}, have become a foundation~\cite{foundationmodel} for modern computer vision~\cite{VTAB,CLIP,ImageNettransferbetter_1,ImageNettransferbetter_2,ImageNettransferbetter_3,betterimageNet,zhaixiaohua,facebookbenchmark}. 
The generalization quality of pre-trained models---whether models are helpful for various downstream tasks---typically determines models' quality. Recent benchmarks for evaluating model generalization mainly collect downstream tasks belonging to two facets: across image domains  (\eg from natural domain to synthetic domain)~\cite{VTAB}; across vision tasks (\eg from image classification to instance segmentation)~\cite{swav,MOCOv1}. In this study, we focus on the third facet: across semantic super-concepts/realms (\eg across pets to scenes).
%
We found that current benchmarks focusing on the last facet either cover a limited range of the semantic realms or are overly cumbersome for evaluation.
%
For example, ImageNet-1k~\cite{russakovsky2015imagenet} only focuses on the mammal, instrumentality, devices, and consumer goods. But these realms struggle to describe the complete realms in the natural domain. Meanwhile, though CLIP~\cite{CLIP} builds a significant benchmark that consists of 24 datasets across a large spectrum of realms, several datasets included have an extensive concept overlapping (\eg CIFAR100~\cite{cifar}, Caltech101~\cite{Caltech101}, and STL-10~\cite{STL10} all have bird 
class) as shown in Fig.~\ref{fig:fig1}(b), resulting in benchmarking on its benchmark suite is less efficient than on an ideal benchmark where datasets included do not conceptually overlap with each other.

%
In this work, we systematically investigates how to build a benchmark for evaluating omni-vision representation that can generalize to a wide range of semantic realms. This benchmark focuses the classification task on the natural image domain.
Through the analysis of ImageNet-1k, we find that the limited number of concepts in its concepts resource, \ie WordNet~\cite{wordnet}, results in its limited realm coverage.\footnote{Annotation budget also limits its realm coverage.} 
Starting from WordNet, we enrich its concepts by integrating new concepts from Wikidata~\cite{wikidata}, building a large ontology with nearly 100,000 concepts. These concepts illustrate a more complete distribution of semantic realms.
Further, we separate these concepts into 21 realm-wise datasets, ensuring these datasets have no overlapping semantic concepts. 
After carefully annotation, we build a benchmark consisting of 7,372 classes and 1,074,346 Creative Commons licenses (CC-BY) data across 21 realms. We illustrated the construction of our benchmark in Fig.~\ref{fig:fig1}, and we refer to it as the Omni-Realm benchmark (\textbf{OmniBenchmark}).\footnote{We use ``Omni" to emphasize the diversity of semantic realms.}
OmniBenchmark has two appealing properties. \textbf{1) Diversity.} the number of realms of OmniBenchmark is twice of ImageNet-1k's, \ie 21 \vs 9 , and the average number of concepts per realm is nine times bigger than ImageNet-1k, \ie 263 \vs 29. \textbf{2) Conciseness.} Since realm-wise datasets have no concept overlapping, OmniBenchmark is concise.

We further investigate how to learn a better omni-vision representation. We purpose an omni-vision representation should not only cluster instances of the \textit{same} concept but also cluster the instances of \textit{related} semantic concept---two concepts within the same realm should imply closer relation than two concepts across the realm, \eg the relation between husky and labrador (two dog species) should be closer than husky and Ferrari 488 (a sports car). 
However, current representative representation learning methods, \eg supervised contrastive learning~\cite{paco,supcon}, commonly construct negative pairs by exhaustive sampling without considering their semantic relation, \eg husky and labrador should have the same possibility of being the negative pair than husky and Ferrari 488.
Motivated by this limitation, we present a novel supervised contrastive learning framework called \textbf{Re}lational \textbf{Co}ntrastive Learning (\textbf{ReCo}). 
ReCo selects the high-quality negative pairs, which belong to different semantic realms, for supervised contrastive learning. ReCo improves the state-of-the-art (SOTA) supervised contrastive learning method (PaCo~\cite{paco}) on the ImageNet-1k by 0.7 points gain.
%

%
We conduct extensive studies on OmniBenchmark to diagnose ReCo and other advances in the omni-vision representation learning including architectures (from CNNs~\cite{resnet} to transformers~\cite{ViT}); learning paradigms (from fully-supervised learning to self-supervised learning); pre-training data volume (from ImageNet-1k~\cite{deng2009imagenet} to IG-1B~\cite{SWSL}). 
We reveal several valuable observations and prove the priority of ReCo: ReCo outperforms PaCo with an average 0.5 points gain on OmniBenchmark.

%
We summarize the contributions of this work as follows.
\begin{itemize}
  \item We propose OmniBenchmark with 21 semantic realm-wise datasets. OmniBenchmark focus on evaluating the concept generalization ability of omni-vision representation thoroughly and efficiently.
  \item We evaluate 22 recent representation learning approaches on OmniBenchmark, uncovering several interesting insights for future research.
  \item A novel supervised contrastive learning method, ReCo, is proposed for encoding the semantic relation information in the supervised contrastive learning framework, achieving competitive performance on ImageNet1k and OmniBenchmark.

\end{itemize}

\section{Related Work}
\noindent\textbf{Representation Learning Methods.}
Representation learning has advanced thanks to improvements in various learning paradigms. 
To avoid the need for supervision, self-supervised leaning~\cite{deepcluster,PIRL,MOCOv1,simclr,swav,Colorization,Colorization_2,instance_1,instance_2,instance_3} has been proposed. 
Recently, Guo \etal~\cite{hcsc} proposed hierarchical contrastive selective coding (HCSC) that improves conventional contrastive learning by implicitly coding the semantic similarity between instances. 
%
Inspired by HCSC, ReCo explicitly encodes the semantic similarity, 
using the label information in the hierarchical concepts structure.
In addition, weakly supervised learning~\cite{SWSL, self-training,self-training_2} focuses on learning from unlabeled data by self-training. The success of representation learning should also owe to architecture designs breakthrough, \eg Vision Transformer~(ViT)~\cite{ViT}. This paper benchmark the most recent methods that have public implementations available. 
Moreover, there have extensive studies that explore to use semantic structure in various ways. Specifically, Bertinetto~\etal~\cite{bertinetto2020making} use the information in the class hierarchy to achieve competitive performance on several datasets. Wang~\etal~\cite{wang2021hierarchical} model the classification process on the semantic hierarchy as a sequential decision-making task. In addition, Wang~\etal~\cite{wang2019deep} propose a deep fuzzy tree model for learning the semantic hierarchy better. Wang~\etal~\cite{wang2021hierarchical2} and Guo~\etal~\cite{guo2022intra} propose very insightful frameworks for leveraging semantic relation information in tasks other than image classification.

\noindent\textbf{Evaluations and Benchmarks.}
The importance of empirical evaluation of representation learning is highlighted by the growing number of major evaluation papers~\cite{betterimageNet,ImageNettransferbetter_1,ImageNettransferbetter_2,ImageNettransferbetter_3,facebookbenchmark}.
%
%
%
CLIP~\cite{CLIP} proposes a significant benchmark that consists of 24 datasets across different semantic realms.
However, CLIP benchmark suite struggles to select the most valuable set of datasets for the benchmark. 
In particular, several datasets have an extensive concept overlapping, making their benchmark suite cumbersome for evaluation.
Visual Decathlon~\cite{decathlon} evaluates the ability of representations to capture simultaneously ten very different visual domains and measures their ability to perform well uniformly. 
Visual Task Adaptation Benchmark (VTAB)~\cite{VTAB} includes datasets from several different domains (natural, specialized, and structured) and annotation information (classification, counting, and \etc) to evaluate methods of varying learning paradigms. 
We argue that the motivation and evaluation protocol of OmniBenchmark are different from VTAB and Visual Decathlon. 
First, the Visual Decathlon studies the multi-task learning ability of models. Its evaluation protocol is train-test. By contrast, OmniBenchmark explores the generalization ability of pre-trained models. Its evaluation protocol is pretrain-train-test.
Secondly, VTAB quantifies the generalization ability of representation to transfer to different image domains and different annotation information (The first and second facet of quantifying representation quality as mentioned in Sec.\ref{sec:introduction}).
OmniBenchmark focuses explicitly on evaluating concept generalization of the classification task in the natural domain. 
Beyond that, ImageNet-COG~\cite{Imagenet-COG} also studies concept generalization. It evaluates the generalization of models from ImageNet-1k to other unseen concepts for ImageNet-1k.
%
%
OmniBenchmark evaluates the generalization of models from any open pre-trained source to extensive categories organized by semantic realms.
\section{The Construction of Omni-Realm Bechmark}
\label{GVB}

In this section, we describe the construction of \Bname. First, we integrate new concepts from Wikidata into WordNet, enlarging the concept storage of WordNet extensively  (Sec.~\ref{subsec:Integrating New Concepts}). For the all concepts of this larger ``WordNet'', we secondly crawl raw images of each of them from Flickr (Sec.~\ref{subsec:Integrating New Concepts}). Thirdly, we select valid concepts from the whole concepts, following multiple steps (Sec.~\ref{subsec:Concept Filtering}). Fourthly we split WordNet into 21 sub-trees, each sub-tree represents a semantic realm and covers a set of valid concepts. (Sec.~\ref{subsec:Realm Selection}) Fifthly, we annotate all the raw images of the valid concepts of these 21 semantic realms (Sec.~\ref{subsec:Image Annotation and deduplication}), forming the 21 realm-wise datasets for benchmarking. In addition, we present detail information of the statistics (Sec.~\ref{subsec:Benchmark Statistics}), and the evaluation protocols of OmniBenchmark. (Sec.~\ref{subsec:Evaluation Protocol}).

\subsection{Integrating New Concepts}
\label{subsec:Integrating New Concepts}
Wikidata~\cite{widerface} contains a large number of concepts, such as different kinds of foods and structures. As the number of concepts in Wikidata continues to grow, we have integrated 170,586 concepts from it so far. These concepts are the leaf nodes in the taxonomy of the Wikidata. 
Referred to~\cite{Tanon2020YAGO4A}, we link Wikidata leaf node concepts to the WordNet through leveraging on the ``sub-classOf'' attribute of Wikidata. 
This attribute denotes the hypernyms relation. 
Further, we obtain raw images for each concept by using the Flickr API.\footnote{https://www.flickr.com/, All the crawled images are strictly followed the CC-BY license.}

\subsection{Concept Filtering}
\label{subsec:Concept Filtering}
After integrating 170,586 new concepts into WordNet, there are nearly 210K concepts in the WordNet. However, not all these concepts are valid for the image classification benchmark. Thus, we manually filter the 210K concepts by the following steps. 
1) We ask annotators to identify and then discard concepts that are related to the offensive content, such as drugs and blood. 
2) Referred to Yang \etal~\cite{yang2020towardsfairer}, we only keep the visual concepts that are highly concrete words.
3) Referred to ImageNet~\cite{russakovsky2015imagenet}, We suggest ensuring the label in our benchmark is mutually exclusive. We thus only keep annotate leaf node in the larger ``WordNet'' created in Sec.~\ref{subsec:Integrating New Concepts}.
4) In our trial annotation, \ie annotation for randomly selected 200 concepts, only 50\% of the crawled data of a specific concept are semantic related to this concept. Since we plan to have 50 images for the test-set of the concept, and 50 images for its train set, we thus discard concepts that have less than 200 crawled data. We illustrate these steps in Fig.~\ref{fig:statistics}~(b).

\subsection{Realm Selection}
\label{subsec:Realm Selection}
\begin{figure}[t]
\centering
\includegraphics[width=\textwidth]{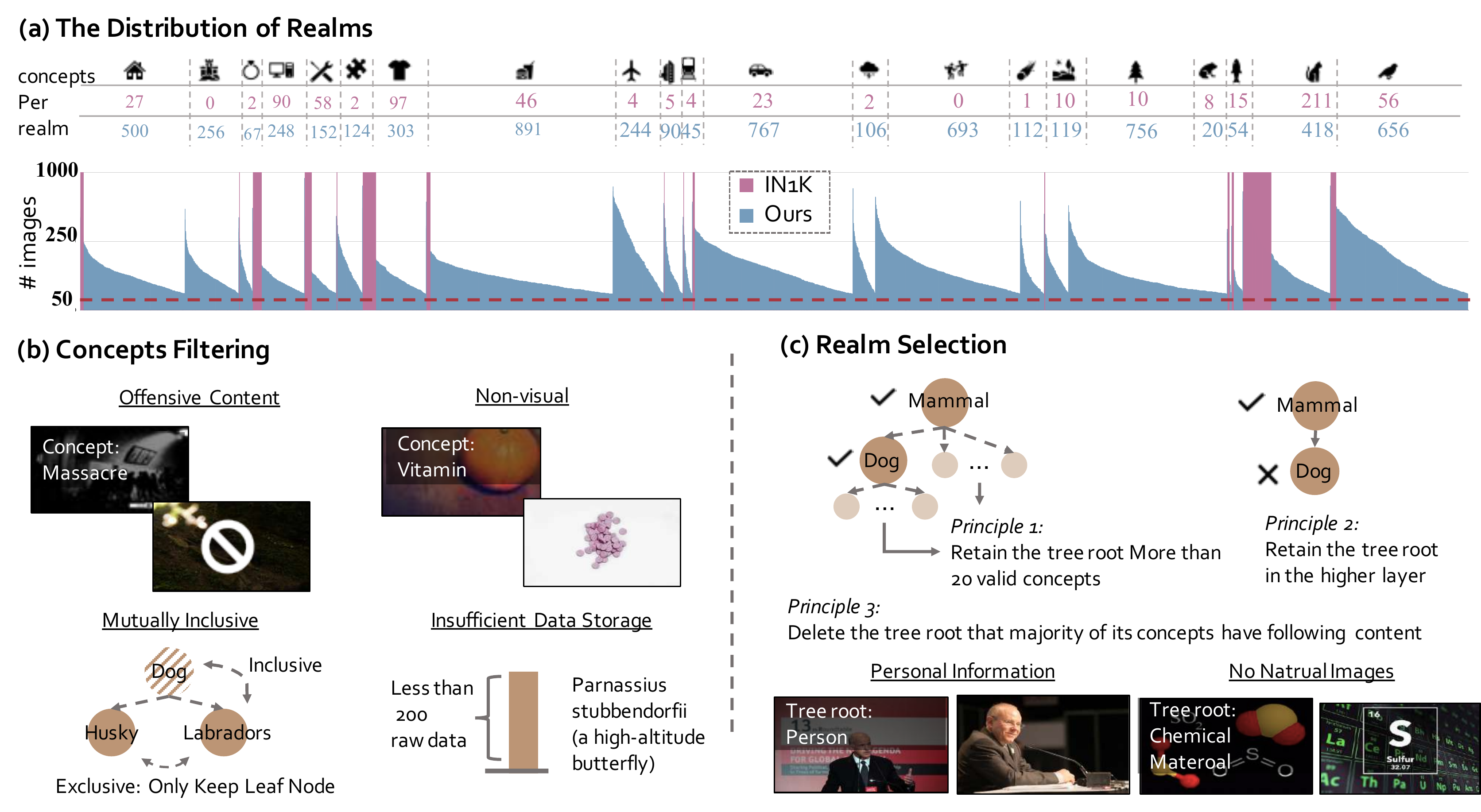}
\caption{\textbf{(a) The statistic of OmniBenchmark}. We compare the distribution of the concepts of each realm in OmniBenchmark and ImageNet-1k. For ImageNet-1k, 12 out of 21 realms are unseen-realm because ImageNet-1k has less than 20 concepts in these realms. We report the number of concepts per realm of  ImageNet-1k and OmniBenchmark in a different color. \textbf{(b) Four principles of filtering concepts.} We discard the concept when it implies offensive content, is a non-visual word, is not a leaf node, and has less than 200 raw data. \textbf{(c) Three principles of selecting realms.} We illustrate three principles to select semantic realms from numerous sub-trees.
}
\label{fig:statistics}
\end{figure}

In WordNet, concepts are organized by the hierarchical structure with many sub-trees.  
We select semantic realms from these sub-trees by following three principles. 1) We select sub-trees that cover at least 20 valid concepts. 2) We discard the sub-tree that is covered by another sub-tree. 3) We discard the sub-tree that most of their concepts imply non-natural images and personal information. For example, we discard the chemical material realm because most of its concepts can only be related to the structural formula image. In addition, we also discard person because its related images contain personal information such as IDs, names. We illustrate these three principles in the Fig.~\ref{fig:statistics}~(c).

\subsection{Image Annotation and De-Duplication}
\label{subsec:Image Annotation and deduplication}
%
For each candidate raw image, we ask five annotators whether this image conforms to its query concept. An image is annotated only if at least 3 out of 5 annotators consider its semantic information closely related to its query concept. More detail information, \eg the annotation interface, is described in the \textit{Supplementary Material}.

To enable a meaningful test of generalization, OmniBenchmark remove duplicates with potential pre-training datasets that includes Bamboo-CLS~\cite{shao2021intern,zhang2022bamboo}, ImageNet-22K~\cite{deng2009imagenet}, PASCAL-VOC~\cite{VOC}, MS-COCO~\cite{COCO} and Places~\cite{places}. 
Specifically, we firstly utilize Difference Hash (DHash)~\cite{DHash} to calculate the hash-code of images of both OmniBenchmark and these pre-training datasets. Then we delete the image that has the same hash-code as any image in these pre-training datasets.

\subsection{Benchmark Statistics}
\label{subsec:Benchmark Statistics}
In Fig.~\ref{fig:statistics}~(a), we compare the distribution of the concepts of each realm of OmniBenchmark and ImageNet-1k. Specifically, 815 concepts of ImageNet-1k are included in these 21 realms, which can fairly reflect the distribution of the ImageNet-1k.\footnote{The other 185 concepts are included in realm that are filtered in Sec.~\ref{subsec:Realm Selection}.}
Fig.~\ref{fig:statistics}~(a) presents that ImageNet-1k covers a very limited number of \textit{ImageNet-1k-seen-realms}, \eg Device, instrumentality consumer\_goods and mammal.\footnote{In the following material, the dataset-seen-realms/dataset-unseen-realms is a set of realms. Each dataset-seen-realm/dataset-unseen-realm in it has at least 20 concepts/fewer than 20 concepts.} Therefore, ImageNet-1k hardly becomes an ideal benchmark to evaluate the generalization ability of the omni-vision representation.  
In contrast, the number of OmniBenchmark-seen-realms is 21 that is twice of ImageNet-1k-seen-realms'. We believe the large spectrum of realms included in OmniBenchmark can relatively thoroughly represents the distribution of realms in the natural domain.

We carefully discuss the potential legal issue of OmniBenchmark, \eg the copyright and privacy issue, in the \textit{supplementary material}. 

\subsection{Evaluation Protocol}
\label{subsec:Evaluation Protocol}
We now present the protocol for OmniBenchmark, and summarize the metrics for the experiments presented in Sec.~\ref{sec:Evaluation models on OmniBenchmark}.

\noindent \textbf{Linear Probing.}
Our benchmark quantifies the generalization ability of visual representation on different semantic realms. We build the evaluation protocol based on the assumption that a good representation is an omni-vision representation that can generalize any specific realm without updating the feature extractor. 
Therefore, we keep the pre-trained backbone frozen and use it as the feature extractor. 
Then, we learn linear logistic regression classifiers for realm-wise datasaets, following the linear probing setting~\cite{facebookbenchmark}.

\noindent \textbf{Metrics.}
We report the top-1 accuracy for all experiments. In addition, to makes the plots clearer and the differences easier to grasp, we follow the metric setting in the ImageNet-CoG~\cite{Imagenet-COG} that plots accuracy relative to the baseline ResNet-50 model pre-trained on ImageNet-1k.

\section{Methodology}

\subsection{Preliminaries}
\noindent\textbf{Self-supervised Contrastive learning.}
A widely-used way to achieve self-supervised learning is contrastive learning, which is proposed in InfoNCE~\cite{InfoMin}. InfoNCE loss works for pulling positive pairs belonging to the same classes closer and push negative pairs belonging to different classes away is an effective way to achieve self-supervised learning.  Specifically, for a set of $N$ sampled pairs, $\left\{x_k,y_k \right\}_{k=1...N}$, the corresponding batch used for training consists of $2N$ pairs, $\left\{\widetilde{x_i},\widetilde{y_i} \right\}_{i=1...2N}$, where $\widetilde{x}_{2k}$ and $\widetilde{x}_{2k-1}$ are two random augmentations of $x_k$ and $\widetilde{y}_{2k} = \widetilde{y}_{2k-1} = y_k$. Finally, a $N$ samples batch forms a $2N$ multiview batch, let $i\in I\equiv \left\{ 1...2N\right\}$ be the index of an augmented sample, and let $j(i)$ be the index of the other augmented sample originating from the same source sample. For a given sample $i$, the loss takes the following form:

\begin{equation}
\mathcal{L}_{i} \!=\! - \log \frac{ \exp({z_{i} \cdot z_{j(i)}}) }{\sum_{a \in A(i)} \exp({z_{i} \cdot z_{a}})}. 
\label{eq:infoNCE} 
\\
\end{equation}
Here, $z_i = Encoder(x_i)$, $A(i) \equiv  I\setminus \left\{ i\right\}$, the index $i$ is called the anchor, index $j(i)$ is called the positive, and the other $2(N-1)$ indexes ($A(i) \equiv  I\setminus \left\{ i\right\}$) are called the negative.

\noindent\textbf{Supervised Contrastive learning.}
To add label information into the self-supervised representation learning, Khosla \etal~\cite{supcon} proposed the supervised contrastive learning (Supcon) in the following way:
\begin{equation}
\mathcal{L}_{i} \!=\!-\sum_{ z_{+} \in P(i)} \log \frac{ \exp({z_{i} \cdot z_{+}}) }{\sum_{z_{k} \in A(i)} \exp({z_{i}\cdot z_{k}})},\ P(i) \equiv \left\{  p \in A(i) : \widetilde{y}_p=\widetilde{y}_i \right\}.
\label{eq:supcon} \\
\end{equation}

\noindent\textbf{Parametric Contrastive learning.}
Cui \etal~\cite{paco} introduce a set of parametric class-wise learnable center $\mathbf{C}= \left\{c_1,c_2,...,c_n\right\}$ into the original supervised contrastive learning, and named a new framework: Parametric Contrastive learning (PaCo). Here, $n$ is the number of the classes. Specifically, the loss is change to.
\begin{equation}
\mathcal{L}_{i} \!=\!-\sum_{ z_{+} \in P(i) \cup \{c_{\widetilde{y}}\} } \!\log \frac{ \exp({z_{i} \cdot z_{+}}) }{\sum_{z_{k} \in A(i) \cup \mathbf{C}} \exp({z_{i} \cdot z_{k}})}.
\label{eq:PaCo}
\end{equation}
PaCo boosts the performance of Supcon on several datasets.

\begin{figure}[t]
\centering
\includegraphics[width=\textwidth]{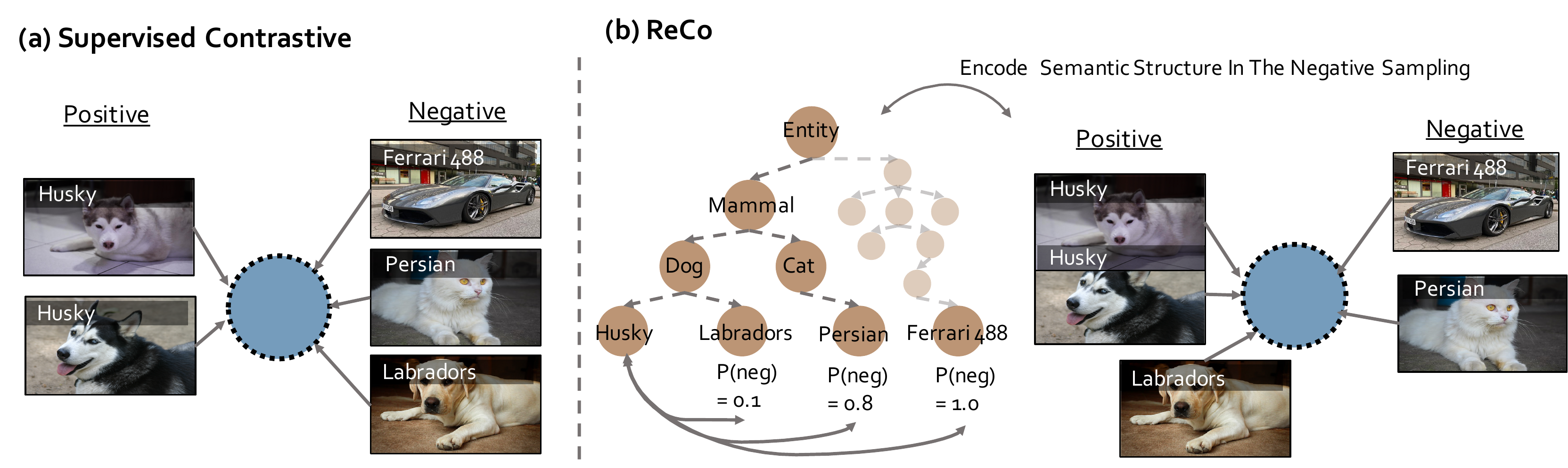}
\caption{\textbf{Supervised contrastive losses \vs ReCo}. The supervised contrastive loss (left, Eq.~\ref{eq:supcon}) contrasts the set of all samples from the same class as positives against the negatives from the remainder of the batch. The ReCo (right, Eq.~\ref{eq:ReCo}), however, contrasts the set of all samples from the same class as positives against the \textit{real semantic dissimilar} negatives from the remainder of the batch. 
Leverage semantic relation when sampling negative samples, ReCo provides an embedding space where elements of similar classes (husky and labrador) are aligned closer than in the supervised case. 
}
\label{fig:ReCo}
\end{figure}

\subsection{Motivation}
Previous Supcon and PaCo methods sample negative samples uniformly over the datasets. However, some negative samples are semantics closely to the query sample, \eg the husky and labrador in the Fig~\ref{fig:ReCo} (a). Wrongly expelling these negative samples from query samples could break the semantic structure to some extent, hampering the representation learning. 

Motivated by this limitation, we aim to carefully select semantic irrelevant negative samples to the query sample instead of selecting negative samples uniformly.
%
In particular, We propose a novel hierarchical instance contrastive learning framework called Relational Contrastive Learning (ReCo). ReCo captures the semantic relation and then transforms this relation into the probability of being sampled as negative samples, as shown in Fig.~\ref{fig:ReCo} (b).

\subsection{Relational Contrastive learning}
The gist of ReCo is to pulling instances of similar semantic classes closer while pushing instances of dissimilar semantic classes far apart. 
For a specific query image $x$, we select real semantic dissimilar negative samples for the contrastive learning by performing Bernoulli sampling on each negative candidate, which is inspired by HCSC~\cite{hcsc}.
To achieve this goal, we first define a similarity $s(m,n)$ measure between two classes $m,n$ as follows. 
\begin{equation}
s(m,n) = - \log \frac{d_{\min}(m,n)+1}{2*\max (l^m,l^n)+1},
\label{eq:distance}
\end{equation}
$m,n$ lie in the hierarchical depth $l^m,l^n$,
and $d_{\min}(m,n)$ denotes the shortest path that connects the $m$ and $n$ in the hypernym/hyponym taxonomy. As shown in Fig~\ref{fig:ReCo} (b), since the dog is the father node of the husky, their shortest distance equals one, and the shortest distance between husky and labrador equals two because they are in sibling relation. 
In addition, $\max (\cdot)$ in the denominator makes the $s(m,q) > s(m,n)$ if $q$ is another class that lies in deeper depth than $n$ and $d_{\min}(m,n) = d_{\min}(m,q)$. We ensure that $m$ should be more related to node that lies in deeper depth because as the hierarchical depth is deeper, the nodes in that depth are more concrete. For example, as shown in Fig~\ref{fig:ReCo} (b), the dog should be semantic closer to the husky than to the mammal.
Moreover, we normalize the similarity between $m$ and other classes for ensuring $s(m,\cdot) \in \left [0,1  \right ]$. 

On such bases, we conduct negative sample selection for a specific instance $x_i$. For the negative candidate $z_{k} \in A(i) $, we are more likely to select it if its $S(\widetilde{y}_i,\widetilde{y}_k)$ is low. The negative sampling follows the Bernoulli sampling:
\begin{equation}
A_{select}(i) = \left\{ \text{\ss} (z_k;P = 1- s(\widetilde{y}_i,\widetilde{y}_k))|z_k \in A(i)\right\},
\label{eq:Bernoulli}
\end{equation}
where $\text{\ss}(z;s)$ denotes a Bernoulli trail of accepting $z$ with similarity $S$. 
By using these refined negative samples selection, we define the
objective function of ReCo as below
\begin{equation}
\mathcal{L}_{i}(ReCo) \!=\!-\sum_{ z_{+} \in P(i)} \log \frac{ \exp({z_{i} \cdot z_{+}}) }{\sum_{z_{k} \in A_{select}(i)} \exp({z_{i}\cdot z_{k}})}. 
\label{eq:ReCo} \\
\end{equation}

In general, ReCo injects the advantages of semantic relation from the hierarchical structure into the negative pairs sampling, and it can effectively regularize the overall contrastive learning:
\begin{equation}
_{f(\theta)}^{\text{min}}\mathcal{L}_{Supcon/PaCo}+\alpha\mathcal{L}_{ReCo}.
\label{eq:overall} \\
\end{equation}
We note that ReCo is not sensitive to the weight $\alpha$. We empirically set $\alpha=1$.


\begin{table*}[t]
\small
\caption{\textbf{Up: Linear classification Result On OmniBenchmark.} We present the Top-1 linear probing accuracy on each realm dataset for 25 models listed in Sec.~\ref{subsec:models}. \textbf{Down: Accuracy relative to the baseline ResNet50.} Accuracy relative to the baseline ResNet50 for the all models, split across the four model categories. For the limited space, we only report the performance of models on 15 out of 21 realms. The complete results are shown in \textit{Supplementary Material}. ImageNet-1k-seen-realms are marked in \underline{underline}. Consumer. denotes consumer\_goods, Locom. denotes locomotive.} 
\label{Tab:performance}
\parbox{\linewidth}{
\centering
\includegraphics[width=\textwidth]{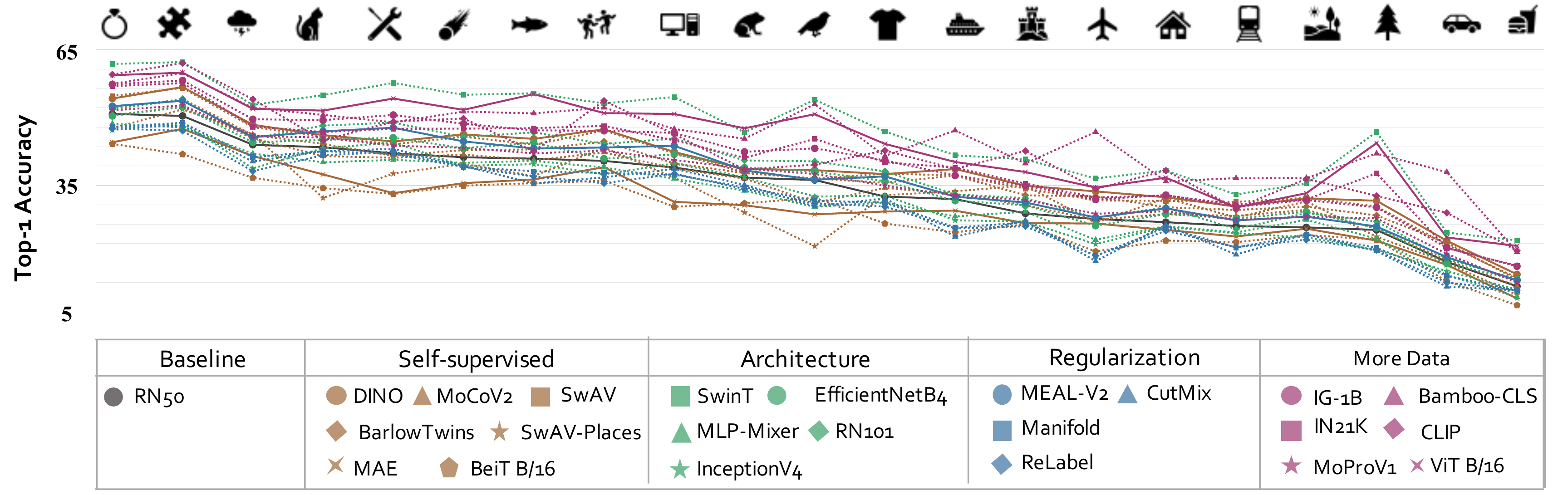}
}
\hfill
\parbox{1.05\linewidth}{
\centering
\tiny
\begin{tabular}{l|llllllllllllllll}
\rotatebox{65}{Method}  &  \rotatebox{65}{Decoration} & 
\rotatebox{65}{\underline{Mammal}} & \rotatebox{65}{\underline{Instrument}} & \rotatebox{65}{Activity} & \rotatebox{65}{\underline{Device}}  & \rotatebox{65}{Bird} & \rotatebox{65}{\underline{Consumer.}} & \rotatebox{65}{Military.}  & \rotatebox{65}{Region}   & \rotatebox{65}{Aircraft} & \rotatebox{65}{\underline{Structure}}   & \rotatebox{65}{Locom.} &
\rotatebox{65}{Plant} & \rotatebox{65}{\underline{Car}} & \rotatebox{65}{\underline{Food}} & \rotatebox{65}{AVG$\uparrow$}  
\\
\Xhline{1.5pt}
DINO~\cite{DINO}           & 3.4   & 2.7       & 2.6       & 7.0  & 3.4  & 2.2   & 5.0  & 6.6  & 6.1  & 6.1  & 5.3  & 4.3  & 6.5  & 4.7  & 2.7  & 4.6  \\
 
MoCov2~\cite{MOCOv2}         & -0.3  & -2.0      & -1.0      & 1.9  & -0.2 & -5.1  & 1.9  & 1.2  & 1.8  & -0.2 & 1.8  & 2.2  & 0.3  & 0.4  & 0.1  & 0.5  \\
 
SwAV~\cite{swav}          & 4.0   & 2.4       & 2.0       & 6.7  & 3.2  & 0.3   & 5.1  & 5.2  & 5.5  & 4.1  & 4.7  & 5.3  & 5.2  & 3.8  & 1.9  & 4.0  \\
 
BarlowTwins~\cite{barlow}    & 1.2   & 0.8       & -0.6      & 4.3  & 0.9  & 0.4   & 3.1  & 5.2  & 4.0  & 5.0  & 3.3  & 3.6  & 3.2  & 4.2  & 1.5  & 2.6  \\
 
SwAV-Places~\cite{contrasting}   & -3.2  & -11.2     & -4.5      & 3.0  & -2.3 & -14.6 & 0.3  & 1.7  & 6.2  & -0.8 & 5.5  & 1.9  & -1.5 & -0.2 & -1.9 & -1.2 \\
 
MAE~\cite{MAE}            & -6.3  & -6.0      & -8.8      & -1.4 & -7.6 & -7.6  & -3.3 & -2.6 & -2.1 & -1.0 & -1.7 & -2.3 & -2.4 & -0.7 & -2.7 & -3.7 \\
 
BeiT B/16~\cite{beit}      & -6.6  & -9.0      & -9.0      & -4.4 & -8.7 & -3.9  & -6.0 & -7.4 & -2.4 & -7.2 & -4.1 & -3.6 & -2.2 & -4.3 & -4.2 & -5.6 \\ \Xhline{1pt}

SwinT~\cite{swin}          & 11.0  & 11.5      & 15.6      & 12.7 & 15.6 & 17.7  & 14.4 & 9.7  & 12.0 & 9.0  & 11.4 & 7.0  & 21.7 & 6.5  & 10.2 & 12.1 \\ 
 
EfficientNetB4~\cite{efficientnet} & -0.5  & 1.6       & 3.5       & 0.5  & 2.9  & 1.9   & 3.4  & -0.4 & 1.8  & -1.5 & 1.8  & 1.6  & 0.5  & -0.4 & 1.5  & 1.5  \\
 
MLP-Mixer~\cite{mlpmixer}      & -2.2  & -3.3      & -1.5      & -1.5 & -2.5 & -5.9  & -1.3 & -3.8 & 0.4  & -4.6 & -0.9 & -1.4 & -2.0 & -2.9 & -2.4 & -2.1 \\
 
ResNet-101~\cite{resnet}      & 1.0   & 4.8       & 6.8       & 3.7  & 6.4  & 4.1   & 5.6  & 1.0  & 3.4  & 0.3  & 3.3  & -0.7 & 1.9  & 0.7  & 1.6  & 3.2  \\
 
InceptionV4~\cite{inceptionv4}   & -3.6  & -0.2      & 2.4       & -3.0 & 0.9  & -3.7  & 0.2  & -4.7 & -1.8 & -5.6 & -1.0 & -1.6 & -4.4 & -2.1 & -0.7 & -1.9 \\ \Xhline{1pt}
 
MEAL-V2~\cite{meal}         & 1.7   & 3.5       & 5.7       & 3.0  & 4.8  & 0.1   & 4.5  & 0.5  & 2.4  & 0.3  & 3.1  & 1.2  & 0.8  & 1.1  & 1.3  & 2.3  \\
 
CutMix~\cite{cutmix}         & -3.1  & -0.8      & 1.0       & -3.6 & -0.3 & -5.8  & -0.4 & -8.3 & -1.6 & -9.3 & -0.6 & -6.3 & -4.7 & -5.5 & -0.9 & -3.2 \\
 
Manifold~\cite{Manifold}       & -2.9  & -0.9      & -0.1      & -2.7 & -1.6 & -4.6  & -1.3 & -6.3 & -2.0 & -8.1 & -1.1 & -4.8 & -4.0 & -3.1 & -1.0 & -2.7 \\
 
ReLabel~\cite{Re-labeling}        & -3.3  & -1.8      & 0.3       & -5.0 & -1.4 & -5.1  & -2.2 & -6.5 & -2.8 & -8.1 & -1.9 & -4.7 & -4.4 & -4.6 & -0.9 & -3.5 \\ \Xhline{1pt}
 
IG-1B~\cite{SWSL}          & 6.5   & 6.0       & 8.5       & 6.7  & 7.7  & 7.0   & 8.0  & 5.2  & 6.2  & 4.3  & 6.0  & 4.2  & 5.0  & 3.1  & 4.4  & 6.1  \\
 
Bamboo-CLS~\cite{zhang2022bamboo}    & 6.7   & 7.2       & 6.9       & 11.8 & 8.5  & 16.8  & 9.0  & 15.2 & 11.1 & 19.2 & 9.0  & 10.6 & 16.9 & 19.9 & 7.6  & 11.1 \\
 
IN21K~\cite{bit}          & 6.1   & 3.4       & 5.4       & 7.7  & 6.0  & 9.1   & 7.6  & 7.0  & 6.6  & 4.8  & 5.9  & 4.8  & 12.5 & 3.1  & 4.6  & 6.2  \\
 
CLIP~\cite{CLIP}           & 8.7   & 1.5       & 7.0       & 13.3 & 7.4  & 3.4   & 10.2 & 6.7  & 13.8 & 6.7  & 11.4 & 3.9  & 7.5  & 10.9 & 7.9  & 7.8  \\
 
MoPro-V1~\cite{MoPro}      & 0.8   & 2.1       & 1.7       & 2.3  & 1.7  & 1.3   & 2.4  & 0.8  & 3.0  & 1.2  & 2.4  & 1.3  & 2.6  & 1.8  & 0.9  & 1.8  \\
 
ViT B/16~\cite{ViT}       & 8.6   & 8.1       & 12.1      & 10.6 & 11.8 & 14.5  & 11.6 & 8.2  & 9.2  & 6.9  & 9.8  & 4.1  & 19.2 & 5.4  & 8.9  & 10.0 \\
\end{tabular}
}
\end{table*}

\section{Systematic Investigation on OmniBenchmark}
\label{sec:Evaluation models on OmniBenchmark}
We now report our thorough experimental studies on OmniBenchmark. These studies evaluate an extensive suite of recent representation learning methods. We first introduce these methods briefly (Sec.~\ref{subsec:models}). Then, we quantify existing intuitions and reveal new insights from the results of the experiments (Sec.~\ref{subsection:Benchmark Results}). Finally, we present the ablation studies of ReCo (Sec.~\ref{subsection:Ablation Study of ReCo}), indicating that ReCo can boost the performance of the state-of-the-art supervised contrastive learning on both the ImageNet-1k and OmniBenchmark.

\subsection{Models}
\label{subsec:models}
We benchmark 22 current models that are split into the following four categories referred to ~\cite{Imagenet-COG}.

\noindent \textbf{Self-supervised.}
We benchmark four types of self-supervised models, including contrastive (MoCov2~\cite{MOCOv2}), clustering-based (SwAV~\cite{swav}, DINO~\cite{DINO}), feature de-correlation (BarlowTwins~\cite{barlow}), masked autoencoder (MAE~\cite{MAE}, BeiT~\cite{beit}) models. These models are based on the ResNet-50 structure.

\noindent \textbf{Architecture.}
We consider several architectures that include CNN based (ResNet-50~\cite{resnet}, ResNet-101~\cite{resnet}, EfficientNet-B4~\cite{efficientnet} and Inception-v4~\cite{inceptionv4}), MLP based (MLP-Mixer~\cite{mlpmixer}), Transformer based (Swin-T~\cite{swin}). All these models are pre-trained on ImageNet-1k~\cite{russakovsky2015imagenet}.

\noindent \textbf{Regularization.}
ResNet-50 sized models with regularization techniques applied during the training phase include distillation (MEAL-V2~\cite{meal}), label augmentation (Manifold-MixUp~\cite{Manifold}, CutMix~\cite{cutmix} and ReLabel~\cite{Re-labeling}).

\noindent \textbf{Larger Scale Data.}
ResNet-50 Model pre-trained on the larger scale data (compared with ImageNet-1k (1M)) includes IG-1B~\cite{SWSL} that is first pre-trained on IG-1B (1000$\times$) and then fine-tuning on ImageNet-1k; CLIP that is image-text models pre-trained on WebImageText (400$\times$); MoProV1~\cite{MoPro} that is pre-trained on the WebVision-V1 (2$\times$); ViT-B/16 that is pre-trained on the ImageNet-22K (14$\times$)~\cite{deng2009imagenet} and then fine-tune on the ImageNet-1k; Bamboo-CLS~\cite{zhang2022bamboo} that is pre-trained on the Bamboo-CLS (65$\times$).

\begin{figure}[t]
\centering
\includegraphics[width=\textwidth]{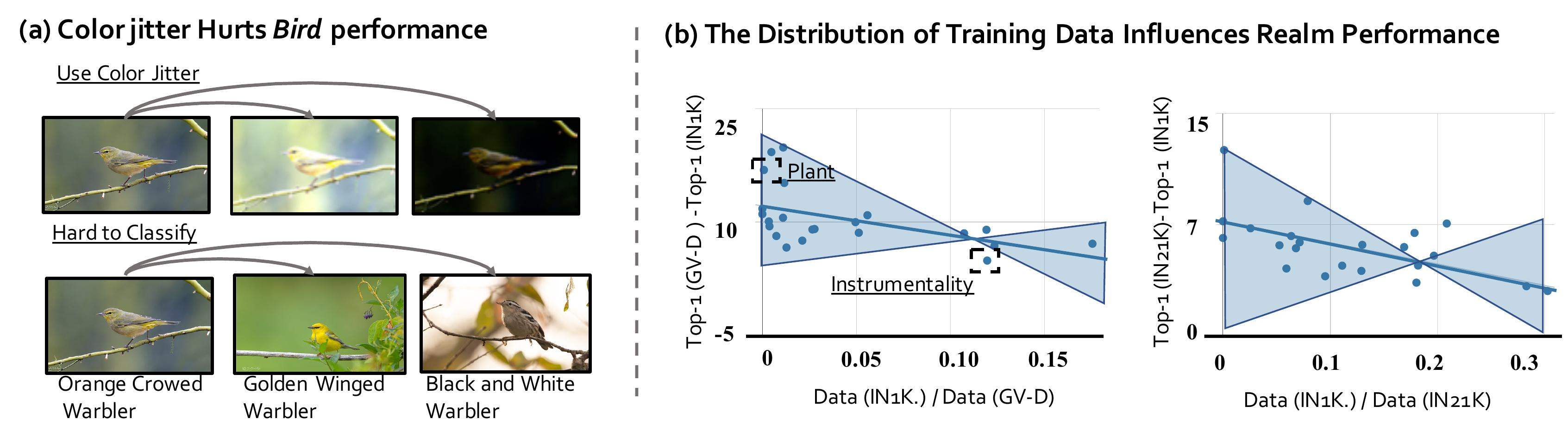}
\caption{\textbf{The Highlighted Observation of Benchmark Results}. (a) Though color jitters make models insensitive to color, it causes the model unable to classify among \textit{Orange Crowed Warbler}, \textit{Golden Winged Warbler} and \textit{Black and White Warbler}.
(b) On the \textit{plant}, the amount of data in IN-1K is 0.08\% of Bamboo-CLS; and on the \textit{instrumentality}, the amount of data in IN-1K is 12.5\% of Bamboo-CLS. Therefore, when it is compared with IN-1K model, the Bamboo-CLS model obtains larger top-1 accuracy gains on the \textit{plant} than on the \textit{instrumentality}. Each points denotes a realm.
}
\label{fig:benchmark result}
\end{figure}

\subsection{Benchmarking Results}
\label{subsection:Benchmark Results}
Table.~\ref{Tab:performance} (up) reports top-1 accuracy of all models on different realms. Table.~\ref{Tab:performance} (down) presents the performance of four model categories relative to the baseline ResNet50. Our main observations are as follows. 

\noindent \textbf{The Selection of Pertaining Data Affects the Realm Performance.} We hypothesize that using a pre-training dataset more similar than ImageNet-1k to a specific end realm dataset will produce a better encoder that performs better on that end realm dataset. We choose one model to test this hypothesis: SwAV-Places. Specifically, SwAV-Places is pre-trained on Places~\cite{places} for 200 epochs based on the SwAV framework, and we download the SwAV-Places from~\cite{contrasting}.  
In Table.~\ref{Tab:performance} (down) Self-supervised split, we find that for realms (\ie structure and region) that are semantic similar to the Places dataset, SwAV-Places achieves the best performance compared with other Self-supervised learning (SSL) methods. This observation justifies our hypothesis.

\noindent \textbf{Strong Augmentation of SSL Hurts Fine-Grained Realm Performance.}
SSL relies on strong data augmentation~\cite{swav,MOCOv2,DINO,barlow} (\eg color jitters) to learn visual representation, encouraging inductive bias. However, some data augmentation strategies may hurt the performance of fine-grain realm~\cite{xiao2021contrastive}, \ie Bird. As shown in Fig.~\ref{Tab:performance} (down) Self-supervised split, except for masked encoder-based methods that use cropping-only augmentation, four out of five SSL methods achieve the worst performance on the bird realm. As illustrated in Fig.~\ref{fig:benchmark result} (a), models that are insensitive to color transformation could struggle to classify between different warbler species.
 
\noindent \textbf{Larger CNN Models Overfit to ImageNet-1k-seen-realms.}
As shown in Tab.~\ref{Tab:performance} (down) Architecture split, we compare ResNet-50 with other CNN models with larger parameters. This figure illustrates that larger CNN models, \ie Inception-v4~\cite{inceptionv4}, EfficientNet-B4~\cite{efficientnet} and ResNet-101~\cite{resnet}, achieve higher performance than ResNet-50 among most of realms. 
However, compared to the ResNet-50, we see larger gains that the these models exhibit on the realm: mammal, instrumentality, device, and consumer\_goods are practically lost for the realm: aircraft, military\_vehicle, and plant. 
We noted that mammal, instrumentality, device, and consumer\_goods are the ImageNet-1k-seen-realm, but aircraft, military\_vehicle, and plant are ImageNet-1k-unseen-realm, we infer that larger CNN models might overfit more to ImageNet-1k-seen-realms.
 

\noindent \textbf{Label Augmentation Techniques are Sensitive to Distribution Shift.}
Robustness to distribution shift is essential to evaluate model generalization ability~\cite{zhou2021domain}. Regularization methods exhibit strong performance gains over ResNet-50 on ImageNet-1k~\cite{Manifold,cutmix,Re-labeling}.
However, on the OmniBenchmark, these methods neither achieve good performance in the ImageNet-1k-unseen-realms nor show performance gain in the ImageNet-1k-seen-realms (\eg instrumentality, consumer\_goods, device), as shown in Table.~\ref{Tab:performance} (down) Regularization split. 
We note that the ImageNet-1k and OmniBenchmark are built at different times, and the distribution shift thus exists in their data. Poor performance of label augmentation methods even on the ImageNet-1k-seen-realms of OmniBenchmark indicates that augmentation techniques are sensitive to the distribution shift. Unlike the larger parameters models that overfit to ImageNet-1k-seen-realm, which performs well on ImageNet-1k-seen-realm but lag in ImageNet-1k-unseen-realm, label augmentation techniques appear to overfit to ImageNet-1k instead of ImageNet-1k-seen-realm.

\noindent \textbf{The Distribution of Training Data Influences Realm Performance.}
As shown in Fig.~\ref{fig:benchmark result} (b), we argue that the distribution of training Data influences realm performance on OmniBenchmark. For example, ImageNet-1k has 1,250 times fewer data than Bamboo-CLS on the plant realm, when compared to ResNet-50, Bamboo-CLS obtains 16.9 points gain on that realm of OmniBenchmark. However, since ImageNet-1k has only 8 times fewer data than Bamboo-CLS on the instrumentality realm, Bamboo-CLS obtains only 6.9 points gain. Each point represents a specific realm.

\noindent \textbf{OmniBenchmark is a Better Indicator for Generalization Ability.} As shown in Bamboo-CLS paper~\cite{zhang2022bamboo}, though DINO outperforms Bamboo-CLS on ImageNet-1k by 5.1\%, Bamboo-CLS outperforms DINO by a large margin on other 9 out of 10 downstream tasks that include Food101~\cite{food}, SUN397~\cite{sun}, OxfordPets~\cite{pets} and \etc, which indicate that the ImageNet-1k performance could not accurately indicates pre-trained models'  generalization ability in most downstream tasks. Surprisingly, Bamboo-CLS shows its better generalization ability than DINO on OmniBenchmark with 6.5 points gain. This observation align with the benchmarking results in Bamboo-CLS paper, and thus indicates that OmniBenchmark is a better indicator than ImageNet-1k for benchmarking pre-trained mode generalization ability.

\begin{table*}[t]
\small
\caption{\textbf{The performance of SupCon, PaCo and ReCo on ImageNet-1k and OmniBenchmark.} On the pre-training dataset, \ie ImageNet-1k, both ReCo ResNet-50 and ResNet-101 models achieve 0.7 top-1 accuracy gain from the state-of-the-art supervised contrastive learning methods, \ie PaCo. On the OmniBenchmark, ReCo ResNet-50 achieves 0.5 top-1 average performance gain from PaCo. Numbers in \textcolor{myred}{red} are the performance gain on the same backbone network.}
\label{Tab:summary_of_dataset}
\centering
\scriptsize
\begin{tabular}{lllll}
Method & Model      & IN1K Top-1 $\uparrow$ & IN1K  Top-5 $\uparrow$ & Omni. AVG$\uparrow$ \\ \Xhline{1.5pt}
Supcon    & ResNet-50  &  75.1         & 92.0   & 35.8       \\
PaCo   & ResNet-50  &    75.9       &   92.5   & 36.4     \\  
ReCo   & ResNet-50  &    \textbf{76.6 \textcolor{myred}{(+0.7)}}       &     \textbf{93.0 \textcolor{myred}{(+0.5)}}  &  \textbf{36.9 \textcolor{myred}{(+0.5)}}   \\ \Xhline{1pt}  
Supcon    & ResNet-101 &    78.9      &    94.4      & 37.9  \\ 
PaCo   & ResNet-101 &     79.1      &    94.4   & 38.4   \\ 
ReCo   & ResNet-101 &     \textbf{79.8 \textcolor{myred}{(+0.7)}}      &    \textbf{94.8 \textcolor{myred}{(+0.4)}} & \textbf{38.7 \textcolor{myred}{(+0.3)}}  
\end{tabular}%
\end{table*}

\subsection{Ablation Study of ReCo}
\label{subsection:Ablation Study of ReCo}
We compare the performance of ReCo, and the state-of-the-art supervised contrastive learning method: parametric contrastive learning (PaCo) on both the pre-training dataset, \ie ImageNet-1k, and OmniBenchmark. Better performance of ReCo indicates that it can effectively boost the current contrastive learning.

\noindent \textbf{Implementation Details.}
Recently, parametric contrastive learning (PaCo)~\cite{paco} achieves better ImageNet-1k performance than the conventional supervised contrastive learning (Supcon). To fair comparison, we reproduce PaCo and train ReCo.
Specifically, we use ResNet-50~\cite{resnet} and ResNet-101~\cite{resnet} as our backbones for experiments. The learning rate decays by a cosine scheduler from 0.1 to 0 with batch size 4096 on 32 GPUs in 200 epochs. These two models are trained using SGD optimizer with momentum $\mu = 0.9$.

\noindent \textbf{Results on IN-1K and OmniBenchmark.}
The experimental results are summarized in Table 7.
Our ResNet-50 model outperforms PaCo baseline models by 0.7\%. And ReCo ResNet-50 improves the average top-1 performance of ReCo on the OmniBenchmark with a 0.5 point gain.

\section{Conclusion}
In our work, we develop a methodology for constructing a large-scale omni-realm benchmark, namely \Bname.
Especially, as the basic building block of the \Bname, we define the visual realm indicated by expert knowledge, \ie WordNet. 
%
Through extensive studies of recent representation learning methods, we find several insights. For example, the distribution of training data influences realm performance
on our OmniBenchamrk. 
%
Besides,  
we propose a novel supervised contrastive learning, \ie ReCo. ReCo selects the semantic dissimilar negative pairs rather than exhaustive sampling, boosting the performance of state-of-the-art supervised contrastive learning methods not only on ImageNet-1k but also on the OmniBenchmark. 
With the advent of the parameter-efficient tuning methods~\cite{jia2022visual,zhou2021learning,zhou2022conditional} in the vision task, we plan to evaluate various representation learning paradigm in the parameter-efficient tuning setting in the future.
Overall,  we hope our work could facilitate future research in omni-vision.
\section*{Acknowledgement}
This work is supported by NTU NAP, MOE AcRF Tier 2 (T2EP20221-0033), and under the RIE2020 Industry Alignment Fund---Industry Collaboration Projects (IAF-ICP) Funding Initiative, as well as cash and in-kind contribution from the industry partner(s). The corresponding author is Jing Shao.

\clearpage
{
\bibliographystyle{splncs04}
\bibliography{eccv2022submission.bbl}
}

\end{document}



\section{Annotation Interface}
\label{Annotation Interface}

\begin{figure}[h]
\centering
\includegraphics[width=\textwidth]{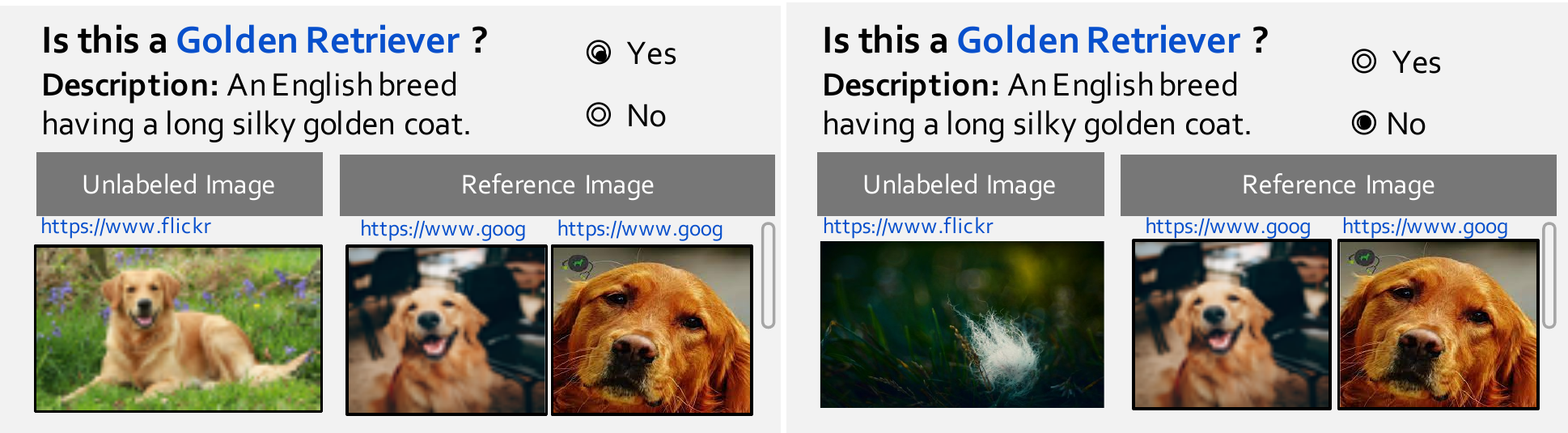}
\caption{\textbf{
The Annotation Interface for OmniBenchmark}. The meta information of each concept consists of a tag, description, and reference images. Annotators determine whether the unlabeled image conforms to the concept. The yes option will be marked if the annotator provides a positive answer to the question above.
}
\label{fig:annotation_interface}
\end{figure}

We illustrate the annotation interface for building OmniBenchmark in Fig.~\ref{fig:annotation_interface}. In addition, the annotator needs to determine the legality of the image. If the unlabeled images contain pornography or any offensive content, annotators should click ``No'' button in this interface.

\section{Experiments Results}
We present the complete Top-1 accuracy (Table~\ref{tab:summary_of_CLS}) and accuracy relative to the baseline ResNet50 (Table~\ref{tab:relative_acc}) of all the 26 models. For all the models, we split them across the four model categories. As shown in Table~\ref{tab:summary_of_CLS}, Swin-T that has similar parameters to ResNet-50 models achieves the best average accuracy on the OmniBenchmark. In addition, GV-D~\cite{shao2021intern} is the best ResNet-50 model, which is pre-trained on the 69M GV-D data.

\begin{table*}[t]
\caption{\textbf{Top-1 accuracy of 26 models on 21 realms.}  We present the complete Top-1 accuracy of 25 models on 21 realms.
ImageNet-1K-seen-realms are marked in \underline{underline}. Consumer. indicates consumer\_goods, Locom. indicates locomotive.}
\tiny
\centering
\ra{1.1}
\label{tab:summary_of_CLS}
\begin{tabular}{l|llllllllllll}
\rotatebox{65}{Method}  &  \rotatebox{65}{Decoration} & 
\rotatebox{65}{\underline{Creation}} & \rotatebox{65}{\underline{Process}} & \rotatebox{65}{Mammal} & \rotatebox{65}{\underline{Instrument.}}  & \rotatebox{65}{Material} & \rotatebox{65}{\underline{Aquatic.}} & \rotatebox{65}{Activity.}  & \rotatebox{65}{Device}   & \rotatebox{65}{Amphibian} & \rotatebox{65}{\underline{Bird}}   &
\\ \Xhline{1.5pt}
RN50~\cite{resnet}           & 50.8 & 50.3 & 44.0 & 43.4 & 42.1 & 41.2 & 40.9 & 40.4 & 38.9 & 36.7 & 36.2 \\
DINO~\cite{DINO}           & 54.2 & 56.7 & 48.4 & 46.1 & 44.7 & 46.3 & 45.3 & 47.4 & 42.4 & 38.6 & 38.4 \\
MoCov2~\cite{MOCOv2}         & 50.5 & 52.4 & 45.4 & 41.5 & 41.1 & 41.8 & 40.5 & 42.3 & 38.7 & 36.5 & 31.1 \\
SwAV~\cite{swav}           & 54.8 & 56.5 & 47.9 & 45.8 & 44.1 & 45.7 & 44.0 & 47.1 & 42.1 & 37.9 & 36.5 \\
BarlowTwins~\cite{barlow}    & 52.0 & 54.1 & 46.1 & 44.2 & 41.5 & 42.8 & 42.9 & 44.7 & 39.9 & 37.7 & 36.5 \\
SwAV-Places~\cite{contrasting}    & 47.6 & 51.8 & 46.3 & 32.2 & 37.6 & 39.8 & 37.0 & 43.4 & 36.7 & 29.0 & 21.6 \\
MAE~\cite{MAE}            & 44.5 & 47.5 & 41.5 & 37.4 & 33.3 & 35.5 & 36.3 & 39.0 & 31.4 & 30.6 & 28.6 \\
BeiT B/16~\cite{beit}      & 44.2 & 41.9 & 36.7 & 34.4 & 33.1 & 34.9 & 35.5 & 36.0 & 30.3 & 31.0 & 32.3 \\ \Xhline{1pt}
EfficientNetB4~\cite{efficientnet} & 50.3 & 52.3 & 44.5 & 45.0 & 45.5 & 43.2 & 44.5 & 40.9 & 41.9 & 38.9 & 38.0 \\
MLP-Mixer~\cite{mlpmixer}      & 48.6 & 47.8 & 42.0 & 40.1 & 40.6 & 40.0 & 40.4 & 38.9 & 36.5 & 33.8 & 30.2 \\
SwinT~\cite{swin}          & 61.8 & 62.2 & 52.8 & 54.9 & 57.6 & 55.0 & 55.3 & 53.1 & 54.5 & 46.7 & 53.9 \\
ResNet-101~\cite{resnet}     & 51.8 & 53.9 & 45.4 & 48.2 & 48.9 & 45.7 & 46.7 & 44.1 & 45.3 & 40.5 & 40.3 \\
InceptionV4~\cite{inceptionv4}    & 47.2 & 48.8 & 38.7 & 43.2 & 44.5 & 39.3 & 39.7 & 37.5 & 39.8 & 36.8 & 32.5 \\ \Xhline{1pt}
MEAL-V2~\cite{meal}        & 52.5 & 53.7 & 45.6 & 46.9 & 47.7 & 44.7 & 43.2 & 43.4 & 43.7 & 38.3 & 36.2 \\
CutMix~\cite{cutmix}         & 47.7 & 48.2 & 40.3 & 42.6 & 43.1 & 39.3 & 35.4 & 36.8 & 38.6 & 35.1 & 30.3 \\
Manifold~\cite{Manifold}       & 47.9 & 48.8 & 41.2 & 42.6 & 41.9 & 39.0 & 38.1 & 37.7 & 37.4 & 34.6 & 31.6 \\
ReLabel~\cite{Re-labeling}        & 47.5 & 47.1 & 38.1 & 41.7 & 42.4 & 39.1 & 37.0 & 35.5 & 37.6 & 34.2 & 31.0 \\ \Xhline{1pt}
IG-1B~\cite{SWSL}          & 57.3 & 58.2 & 49.6 & 49.4 & 50.5 & 48.6 & 47.2 & 47.1 & 46.6 & 42.4 & 43.2 \\
GV-D~\cite{shao2021intern}      & 57.4 & 59.8 & 52.1 & 50.6 & 48.9 & 51.3 & 50.9 & 52.2 & 47.5 & 45.3 & 52.9 \\
IN21K~\cite{bit}          & 56.9 & 57.6 & 48.0 & 46.8 & 47.5 & 47.5 & 47.6 & 48.1 & 44.9 & 41.1 & 45.3 \\
CLIP~\cite{CLIP}           & 59.5 & 62.0 & 54.0 & 44.9 & 49.1 & 49.8 & 42.5 & 53.7 & 46.3 & 38.5 & 39.6 \\
MoPro-V1~\cite{MoPro}       & 51.6 & 52.7 & 45.9 & 45.6 & 43.8 & 43.3 & 42.1 & 42.7 & 40.6 & 38.7 & 37.5 \\
ViT B/16~\cite{ViT}       & 59.3 & 59.9 & 51.9 & 51.5 & 54.2 & 51.7 & 55.2 & 51.0 & 50.8 & 47.6 & 50.7 \\ \Xhline{1pt}
Peco RN50      & 51.4 & 53.5 & 45.0 & 45.2 & 43.4 & 43.6 & 43.5 & 43.1 & 41.3 & 39.3 & 38.8 \\
Reco RN50      & 52.5 & 53.9 & 45.8 & 45.3 & 43.9 & 44.4 & 44.3 & 43.1 & 41.6 & 40.1 & 38.7 \\
Peco RN101     & 54.0 & 54.8 & 47.3 & 48.0 & 46.5 & 45.7 & 46.0 & 45.0 & 43.6 & 39.4 & 41.1 \\
Reco RN101     & 54.3 & 55.5 & 46.7 & 48.2 & 47.4 & 46.4 & 45.4 & 45.6 & 44.7 & 41.5 & 40.5 \\ \Xhline{1pt}
\rotatebox{65}{Method}  &  \rotatebox{65}{Consumer.} & 
\rotatebox{65}{\underline{Military.}} & \rotatebox{65}{\underline{Region}} & \rotatebox{65}{Aircraft} & \rotatebox{65}{\underline{Structure}}  & \rotatebox{65}{Locomot.} & \rotatebox{65}{\underline{Geologi.}} & \rotatebox{65}{Plant.}  & \rotatebox{65}{Car}   & \rotatebox{65}{Food}  & \rotatebox{65}{AVG$\uparrow$}  
\\ \Xhline{1.5pt}
RN50~\cite{resnet}           & 32.5 & 32.0 & 28.8 & 27.6 & 26.9 & 26.0 & 25.7 & 25.1 & 18.1 & 12.7 & 34.3 \\
DINO~\cite{DINO}           & 37.5 & 38.6 & 34.9 & 33.7 & 32.4 & 30.3 & 32.1 & 31.6 & 22.7 & 15.3 & 38.9 \\
MoCov2~\cite{MOCOv2}         & 34.4 & 33.1 & 30.6 & 27.4 & 28.7 & 28.2 & 29.6 & 25.4 & 18.5 & 12.8 & 34.8 \\
SwAV~\cite{swav}           & 37.6 & 37.2 & 34.3 & 31.7 & 31.6 & 31.3 & 32.2 & 30.3 & 21.9 & 14.6 & 38.3 \\
BarlowTwins~\cite{barlow}    & 35.6 & 37.2 & 32.8 & 32.6 & 30.2 & 29.6 & 30.3 & 28.4 & 22.3 & 14.2 & 36.9 \\
SwAV-Places~\cite{swav}            & 32.8 & 33.7 & 34.9 & 26.7 & 32.2 & 27.9 & 31.5 & 23.7 & 17.9 & 10.8 & 33.1 \\
MAE~\cite{MAE}      & 29.2 & 29.4 & 26.7 & 26.6 & 25.1 & 23.7 & 25.4 & 22.8 & 17.4 & 10.0 & 30.6 \\ 
BeiT B/16~\cite{beit} & 26.5 & 24.6 & 26.4 & 20.4 & 22.8 & 22.4 & 24.0 & 23.0 & 13.8 & 8.5  & 30.1 \\ \Xhline{1pt}
EfficientNetB4~\cite{efficientnet}      & 35.9 & 31.6 & 30.6 & 26.1 & 28.7 & 27.6 & 29.2 & 25.6 & 17.7 & 14.2 & 35.8 \\
MLP-Mixer~\cite{mlpmixer}          & 31.2 & 28.1 & 29.2 & 23.0 & 26.0 & 24.6 & 27.4 & 23.2 & 15.2 & 10.3 & 32.2 \\
SwinT~\cite{swin}     & 46.9 & 41.6 & 40.7 & 36.5 & 38.3 & 33.0 & 35.4 & 46.8 & 24.5 & 22.8 & 46.4 \\
ResNet-101~\cite{resnet}   & 38.1 & 33.0 & 32.2 & 27.8 & 30.1 & 25.3 & 28.9 & 27.0 & 18.7 & 14.2 & 37.4 \\ 
InceptionV4~\cite{inceptionv4}        & 32.8 & 27.3 & 27.0 & 22.0 & 25.8 & 24.4 & 23.5 & 20.8 & 16.0 & 11.9 & 32.3 \\ \Xhline{1pt}
MEAL-V2~\cite{meal}         & 37.0 & 32.5 & 31.2 & 27.9 & 29.9 & 27.2 & 28.1 & 25.9 & 19.1 & 14.0 & 36.6 \\
 CutMix~\cite{cutmix}     & 32.1 & 23.6 & 27.2 & 18.2 & 26.3 & 19.7 & 24.3 & 20.5 & 12.6 & 11.8 & 31.1 \\
Manifold~\cite{Manifold}         & 31.2 & 25.7 & 26.8 & 19.4 & 25.7 & 21.2 & 24.1 & 21.2 & 15.0 & 11.7 & 31.6 \\ 
ReLabel~\cite{Re-labeling}            & 30.3 & 25.5 & 25.9 & 19.5 & 24.9 & 21.3 & 23.0 & 20.7 & 13.5 & 11.8 & 30.8 \\\Xhline{1pt}
IG-1B~\cite{SWSL}      & 40.5 & 37.2 & 34.9 & 31.9 & 32.8 & 30.2 & 31.8 & 30.1 & 21.2 & 17.1 & 40.4 \\
GV-D~\cite{shao2021intern}          & 41.5 & 47.1 & 39.8 & 46.8 & 35.9 & 36.6 & 36.6 & 42.0 & 37.9 & 20.2 & 45.4 \\
IN21K~\cite{bit}         & 40.1 & 38.9 & 35.4 & 32.3 & 32.8 & 30.8 & 31.9 & 37.7 & 21.1 & 17.2 & 40.4 \\
CLIP~\cite{CLIP}       & 42.7 & 38.6 & 42.6 & 34.2 & 38.2 & 29.9 & 36.4 & 32.6 & 28.9 & 20.5 & 42.1 \\
MoPro-V1~\cite{MoPro}       & 34.9 & 32.8 & 31.8 & 28.7 & 29.3 & 27.3 & 28.1 & 27.7 & 19.9 & 13.5 & 36.1 \\ 
ViT B/16~\cite{ViT} & 44.2 & 40.2 & 37.9 & 34.5 & 36.7 & 30.1 & 33.3 & 44.3 & 23.5 & 21.6 & 45.8 \\\Xhline{1pt}
Peco RN50       & 35.1 & 34.9 & 31.0 & 30.6 & 28.9 & 28.0 & 28.1 & 27.4 & 20.6 & 13.8 & 36.4 \\
Reco RN50      & 35.5 & 35.1 & 31.4 & 31.1 & 29.1 & 28.6 & 28.6 & 27.8 & 20.6 & 14.1 & 36.9 \\
Peco RN101    & 37.8 & 37.4 & 32.8 & 33.3 & 30.9 & 28.7 & 29.3 & 28.7 & 22.1 & 15.2 & 38.5 \\
Reco RN101     & 38.4 & 36.9 & 32.7 & 33.8 & 31.0 & 28.4 & 28.9 & 28.6 & 21.8 & 15.5 & 38.7
\end{tabular}
\end{table*}

\begin{table*}[t]
\caption{\textbf{Accuracy of 26 models on 21 realms relative to the baseline ResNet50.} 
We present the complete relative accuracy of 25 models on 21 realms.
ImageNet-1K-seen-realms are marked in \underline{underline}. Consumer. indicates consumer\_goods, Locom. indicates locomotive..}
\tiny
\centering
\ra{1.1}
\label{tab:relative_acc}
\begin{tabular}{l|llllllllllll}
\rotatebox{65}{Method}  &  \rotatebox{65}{Decoration} & 
\rotatebox{65}{\underline{Creation}} & \rotatebox{65}{\underline{Process}} & \rotatebox{65}{Mammal} & \rotatebox{65}{\underline{Instrument.}}  & \rotatebox{65}{Material} & \rotatebox{65}{\underline{Aquatic.}} & \rotatebox{65}{Activity.}  & \rotatebox{65}{Device}   & \rotatebox{65}{Amphibian} & \rotatebox{65}{\underline{Bird}}   &
\\ \Xhline{1.5pt}
DINO~\cite{DINO}           & 3.4             & 6.4               & 4.4     & 2.7      & 2.6             & 5.1        & 4.4                   & 7.0      & 3.4    & 1.9       & 2.2     \\
MoCov2~\cite{MOCOv2}         & -0.3            & 2.1               & 1.4     & -2.0     & -1.0            & 0.6        & -0.4                  & 1.9      & -0.2   & -0.2      & -5.1    \\
SwAV~\cite{swav} & 4.0             & 6.2               & 3.9     & 2.4      & 2.0             & 4.5        & 3.1                   & 6.7      & 3.2    & 1.2       & 0.3     \\
BarlowTwins~\cite{barlow}    & 1.2             & 3.7               & 2.2     & 0.8      & -0.6            & 1.6        & 2.0                   & 4.3      & 0.9    & 1.0       & 0.4     \\
SwAV-Places~\cite{contrasting}    & -3.2            & 1.5               & 2.4     & -11.2    & -4.5            & -1.5       & -3.9                  & 3.0      & -2.3   & -7.7      & -14.6   \\
MAE~\cite{MAE}           & -6.3            & -2.9              & -2.5    & -6.0     & -8.8            & -5.7       & -4.6                  & -1.4     & -7.6   & -6.1      & -7.6    \\
BeiT B/16~\cite{beit}      & -6.6            & -8.5              & -7.3    & -9.0     & -9.0            & -6.3       & -5.4                  & -4.4     & -8.7   & -5.7      & -3.9    \\ \Xhline{1pt}
EfficientNetB4~\cite{efficientnet} & -0.5            & 2.0               & 0.5     & 1.6      & 3.5             & 2.0        & 3.6                   & 0.5      & 2.9    & 2.2       & 1.9     \\
MLP-Mixer~\cite{mlpmixer}      & -2.2            & -2.6              & -2.0    & -3.3     & -1.5            & -1.2       & -0.5                  & -1.5     & -2.5   & -2.9      & -5.9    \\
SwinT~\cite{swin}          & 11.0            & 11.9              & 8.8     & 11.5     & 15.6            & 13.8       & 14.4                  & 12.7     & 15.6   & 10.0      & 17.7    \\
ResNet-101~\cite{resnet}     & 1.0             & 3.6               & 1.5     & 4.8      & 6.8             & 4.5        & 5.8                   & 3.7      & 6.4    & 3.8       & 4.1     \\
InceptionV4~\cite{inceptionv4}    & -3.6            & -1.6              & -5.2    & -0.2     & 2.4             & -1.9       & -1.2                  & -3.0     & 0.9    & 0.1       & -3.7    \\ \Xhline{1pt}
MEAL-V2~\cite{meal}        & 1.7             & 3.3               & 1.7     & 3.5      & 5.7             & 3.5        & 2.3                   & 3.0      & 4.8    & 1.6       & 0.1     \\
CutMix~\cite{cutmix}         & -3.1            & -2.2              & -3.7    & -0.8     & 1.0             & -1.9       & -5.5                  & -3.6     & -0.3   & -1.6      & -5.8    \\
Manifold~\cite{Manifold}       & -2.9            & -1.5              & -2.7    & -0.9     & -0.1            & -2.2       & -2.8                  & -2.7     & -1.6   & -2.1      & -4.6    \\
ReLabel~\cite{Re-labeling}        & -3.3            & -3.3              & -5.8    & -1.8     & 0.3             & -2.2       & -3.9                  & -5.0     & -1.4   & -2.5      & -5.1    \\ \Xhline{1pt}
IG-1B~\cite{SWSL}          & 6.5             & 7.8               & 5.6     & 6.0      & 8.5             & 7.4        & 6.3                   & 6.7      & 7.7    & 5.7       & 7.0     \\
GV-D~\cite{shao2021intern}      & 6.7             & 9.5               & 8.2     & 7.2      & 6.9             & 10.1       & 10.0                  & 11.8     & 8.5    & 8.6       & 16.8    \\
IN21K~\cite{bit}          & 6.1             & 7.3               & 4.0     & 3.4      & 5.4             & 6.3        & 6.7                   & 7.7      & 6.0    & 4.4       & 9.1     \\
CLIP~\cite{CLIP}           & 8.7             & 11.7              & 10.1    & 1.5      & 7.0             & 8.6        & 1.6                   & 13.3     & 7.4    & 1.8       & 3.4     \\
MoPro-V1~\cite{MoPro}       & 0.8             & 2.3               & 1.9     & 2.1      & 1.7             & 2.1        & 1.2                   & 2.3      & 1.7    & 2.0       & 1.3     \\
ViT B/16~\cite{ViT}       & 8.6             & 9.6               & 8.0     & 8.1      & 12.1            & 10.5       & 14.3                  & 10.6     & 11.8   & 10.9      & 14.5    \\ \Xhline{1pt}
PeCo RN50      & 0.6  & 3.2  & 1.0  & 1.8  & 1.3  & 2.4  & 2.6  & 2.7  & 2.3  & 2.6  & 2.7  \\
ReCo RN50      & 1.7  & 3.6  & 1.9  & 1.9  & 1.8  & 3.2  & 3.4  & 2.7  & 2.7  & 3.4  & 2.5  \\
PeCo RN101     & 3.3  & 4.5  & 3.3  & 4.6  & 4.4  & 4.5  & 5.2  & 4.6  & 4.7  & 2.7  & 5.0  \\
ReCo RN101     & 3.5  & 5.2  & 2.7  & 4.8  & 5.3  & 5.2  & 4.5  & 5.2  & 5.7  & 4.8  & 4.4  \\
\Xhline{1pt}
\rotatebox{65}{Method}  &  \rotatebox{65}{Consumer.} & 
\rotatebox{65}{\underline{Military.}} & \rotatebox{65}{\underline{Region}} & \rotatebox{65}{Aircraft} & \rotatebox{65}{\underline{Structure}}  & \rotatebox{65}{Locomot.} & \rotatebox{65}{\underline{Geologi.}} & \rotatebox{65}{Plant.}  & \rotatebox{65}{Car}   & \rotatebox{65}{Food}  & \rotatebox{65}{AVG$\uparrow$}  
\\ \Xhline{1.5pt}
DINO~\cite{DINO}           & 5.0             & 6.6               & 6.1     & 6.1      & 5.5             & 4.3        & 6.4                   & 6.5      & 4.7    & 2.7       & 4.6     \\
MoCov2~\cite{MOCOv2}         & 1.9             & 1.2               & 1.8     & -0.2     & 1.8             & 2.2        & 3.9                   & 0.3      & 0.4    & 0.1       & 0.5     \\
SwAV~\cite{swav}           & 5.1             & 5.2               & 5.5     & 4.1      & 4.7             & 5.3        & 6.5                   & 5.2      & 3.8    & 1.9       & 4.0     \\
BarlowTwins~\cite{barlow}    & 3.1             & 5.2               & 4.0     & 5.0      & 3.3             & 3.6        & 4.6                   & 3.2      & 4.2    & 1.5       & 2.6     \\
SwAV-Places~\cite{swav}    & 0.3             & 1.7               & 6.2     & -0.8     & 5.3             & 1.9        & 5.8                   & -1.5     & -0.2   & -1.9      & -1.2    \\
MAE~\cite{MAE}            & -3.3            & -2.6              & -2.1    & -1.0     & -1.7            & -2.3       & -0.3                  & -2.4     & -0.7   & -2.7      & -3.7    \\
BeiT B/16~\cite{beit}      & -6.0            & -7.4              & -2.4    & -7.2     & -4.1            & -3.6       & -1.7                  & -2.2     & -4.3   & -4.2      & -5.6    \\ \Xhline{1pt}
EfficientNetB4~\cite{efficientnet} & 3.4             & -0.4              & 1.8     & -1.5     & 1.8             & 1.6        & 3.5                   & 0.5      & -0.4   & 1.5       & 1.5     \\
MLP-Mixer~\cite{mlpmixer}      & -1.3            & -3.8              & 0.4     & -4.6     & -0.9            & -1.4       & 1.7                   & -2.0     & -2.9   & -2.4      & -2.1    \\
SwinT~\cite{swin}          & 14.4            & 9.7               & 12.0    & 9.0      & 11.4            & 7.0        & 9.7                   & 21.7     & 6.5    & 10.2      & 12.1    \\
ResNet-101~\cite{resnet}     & 5.6             & 1.0               & 3.4     & 0.3      & 3.3             & -0.7       & 3.2                   & 1.9      & 0.7    & 1.6       & 3.2     \\
InceptionV4~\cite{inceptionv4}    & 0.2             & -4.7              & -1.8    & -5.6     & -1.0            & -1.6       & -2.2                  & -4.4     & -2.1   & -0.7      & -1.9    \\ \Xhline{1pt}
MEAL-V2~\cite{meal}        & 4.5             & 0.5               & 2.4     & 0.3      & 3.1             & 1.2        & 2.4                   & 0.8      & 1.1    & 1.3       & 2.3     \\
CutMix~\cite{cutmix}         & -0.4            & -8.3              & -1.6    & -9.3     & -0.6            & -6.3       & -1.5                  & -4.7     & -5.5   & -0.9      & -3.2    \\
Manifold~\cite{Manifold}       & -1.3            & -6.3              & -2.0    & -8.1     & -1.1            & -4.8       & -1.6                  & -4.0     & -3.1   & -1.0      & -2.7    \\
ReLabel~\cite{Re-labeling}        & -2.2            & -6.5              & -2.8    & -8.1     & -1.9            & -4.7       & -2.7                  & -4.4     & -4.6   & -0.9      & -3.5    \\ \Xhline{1pt}
IG-1B~\cite{SWSL}          & 8.0             & 5.2               & 6.2     & 4.3      & 6.0             & 4.2        & 6.1                   & 5.0      & 3.1    & 4.4       & 6.1     \\
GV-D~\cite{shao2021intern}      & 9.0             & 15.2              & 11.1    & 19.2     & 9.0             & 10.6       & 10.9                  & 16.9     & 19.9   & 7.6       & 11.1    \\
IN21K~\cite{bit}          & 7.6             & 7.0               & 6.6     & 4.8      & 5.9             & 4.8        & 6.2                   & 12.5     & 3.1    & 4.6       & 6.2     \\
CLIP~\cite{CLIP}           & 10.2            & 6.7               & 13.8    & 6.7      & 11.4            & 3.9        & 10.7                  & 7.5      & 10.9   & 7.9       & 7.8     \\
MoPro-V1~\cite{MoPro}       & 2.4             & 0.8               & 3.0     & 1.2      & 2.4             & 1.3        & 2.4                   & 2.6      & 1.8    & 0.9       & 1.8     \\
ViT B/16~\cite{ViT}       & 11.6            & 8.2               & 9.2     & 6.9      & 9.8             & 4.1        & 7.6                   & 19.2     & 5.4    & 8.9       & 10.0  \\ \Xhline{1pt}
PeCo RN50      & 2.6  & 2.9  & 2.2  & 3.0  & 2.1  & 2.0  & 2.4  & 2.2  & 2.5  & 1.1  & 2.1  \\
ReCo RN50      & 3.0  & 3.1  & 2.6  & 3.5  & 2.3  & 2.6  & 2.9  & 2.6  & 2.6  & 1.4  & 2.6  \\
PeCo RN101     & 5.3  & 5.5  & 4.0  & 5.8  & 4.0  & 2.7  & 3.6  & 3.6  & 4.0  & 2.5  & 4.2  \\
ReCo RN101     & 5.9  & 4.9  & 4.0  & 6.3  & 4.1  & 2.4  & 3.2  & 3.5  & 3.8  & 2.9  & 4.4  \\
\end{tabular}
\end{table*} 

\section{Social Impact}
\label{sec:Data_Overlap_Analysis}

The usage of OmniBenchmark might bring several risks, such as privacy, and problematic content. We
discuss these risks and their mitigation strategies as follows.

\noindent \textbf{Copyright.} All images in this paper and the dataset are licensed by the CC-BY (https://creativecommons. org/licenses/by/2.0/) license and contain information on their creator. On the above url, this license is described as follows:

\begin{itemize}
 \item Copy and redistribute the material in any medium or format. 
 \item Remix, transform, and build upon the material for any purpose, even commercially. 
 \item This license is acceptable for Free Cultural Works. 
 \item The licensor cannot revoke these freedoms as long as you follow the license terms. 
\end{itemize}

Referred to LAION-400M~\cite{laion400m}, Conceptual 12M~\cite{cc12m} and MS-COCO~\cite{COCO}, we only present the lists of URLs to this data. We build the meta file as follow.
\begin{equation}
\text{[image\_url]\ [class\_index]} 
\nonumber
\end{equation}

\noindent \textbf{Problematic Content.} The problematic contents such as drugs, nudity, and other offensive content exist in the web data. As mentioned in Sec.~\ref{Annotation Interface}, the annotators were asked to discard such images instead of conducting annotation.

\noindent \textbf{Privacy.} To mitigate privacy issues with public visual
datasets, researchers have attempted to obfuscate private information before publishing the data~\cite{privacy1,privacy2}. We plan to follow this line of work to blur faces, license plates in our new annotated data.
In addition, if the original picture found at the URL present on the OmniBenchmark on the record states users' names, phone numbers, or any personal information, users can request a takedown of this image.

\noindent \textbf{Bias.} The images were crawled from Flickr thus inheriting all the biases of that website. The usage of user generated data might bring the risk of bias. We plan to tackle this problem by balancing various categories.